\documentclass[twoside,11pt]{article}

\usepackage[cachedir=.,frozencache]{minted}
\usepackage[utf8]{inputenc} 
\usepackage{graphicx}
\usepackage{caption}
\usepackage{subcaption}
\usepackage{stmaryrd}
\usepackage[colorinlistoftodos]{todonotes}
\usepackage[abbrvbib]{jmlr2e} 

\usepackage{xspace}
\newcommand{\MATLAB}{\textsc{Matlab}\xspace}

\graphicspath{{images/}}

\usepackage{lastpage}
\jmlrheading{21}{2020}{1-\pageref{LastPage}}{8/19; Revised
7/20}{7/20}{19-678}{William de Vazelhes and CJ Carey and Yuan Tang and Nathalie Vauquier and Aur\'elien Bellet}
\ShortHeadings{metric-learn: Metric Learning Algorithms in Python}{de Vazelhes, Carey, Tang, Vauquier and Bellet}

\firstpageno{1}

\begin{document}

\title{metric-learn: Metric Learning Algorithms in Python}

\author{\name William de Vazelhes\thanks{Most of the work was carried out while the author was affiliated with INRIA, France.} \email wdevazelhes@gmail.com \\
       \addr Paris Research Center, Huawei Technologies\\
       92100 Boulogne-Billancourt, France
       \AND
       \name CJ Carey \email perimosocordiae@gmail.com \\
       \addr Google LLC \\
       111 8th Ave, New York, NY 10011, USA
       \AND
       \name Yuan Tang \email terrytangyuan@gmail.com \\
       \addr Ant Group \\ 
       525 Almanor Ave, Sunnyvale, CA 94085, USA
       \AND
       \name Nathalie Vauquier \email nathalie.vauquier@inria.fr \\
       \name Aur\'elien Bellet \email aurelien.bellet@inria.fr \\
       \addr Magnet Team, INRIA Lille – Nord Europe\\
       59650 Villeneuve d’Ascq, France
       }

\editor{Balazs Kegl}

\maketitle

\begin{abstract}
\texttt{metric-learn} is an open source Python package implementing supervised and weakly-supervised distance metric learning algorithms. As part of \texttt{scikit-learn-contrib}, it provides a unified interface compatible with \texttt{scikit-learn} which allows to easily perform cross-validation, model selection, and pipelining with other machine learning estimators. \texttt{metric-learn} is thoroughly tested and available on PyPi under the MIT license.
\end{abstract}


\section{Introduction}

Many approaches in machine learning require a measure of distance between data
points. Traditionally, practitioners would choose a standard distance metric
(Euclidean, City-Block, Cosine, etc.) using a priori knowledge of the
domain. However, it is often difficult to design metrics that are well-suited
to the particular data and task of interest.
Distance metric learning, or simply metric learning \citep{Bellet15}, aims at
automatically constructing task-specific distance metrics from data. A key advantage of metric learning is that it can be applied beyond the standard supervised learning setting (data points associated with labels), in situations where only weaker forms of supervision are available (e.g., pairs of points that should be similar/dissimilar). The learned distance metric can be used to perform retrieval tasks such as finding elements (images, documents) of a database that are semantically closest to a query element. It can also be plugged into other machine learning algorithms, for instance to improve the accuracy of nearest neighbors models (for classification, regression, anomaly detection...) or to bias the clusters found by clustering algorithms towards the intended semantics. Finally, metric learning can be used to perform dimensionality reduction.
These use-cases highlight the importance of integrating metric learning with the rest of the machine learning pipeline and tools.

\texttt{metric-learn} is an open source package for metric learning in Python, which implements many popular metric-learning algorithms with different levels of supervision through a unified interface.
Its API is compatible with \texttt{scikit-learn} \citep{scikit-learn}, a prominent machine learning library in Python. This allows for streamlined model selection, evaluation, and pipelining with other estimators.

\paragraph{Positioning with respect to other packages.} Many metric learning algorithms were originally implemented by their authors in \MATLAB without a common API convention.\footnote{See \url{https://www.cs.cmu.edu/~liuy/distlearn.htm} for a list of \MATLAB implementations.}  In R, the package \texttt{dml} \citep{Tang18} implements several metric learning algorithms with a unified interface but is not tightly integrated with any general-purpose machine learning library. In Python, \texttt{pyDML} \citep{pyDML} contains mainly fully supervised and unsupervised algorithms, while \texttt{pytorch-metric-learning}\footnote{\url{http://github.com/KevinMusgrave/pytorch-metric-learning}} focuses on deep metric learning using the \texttt{pytorch} framework \citep{pytorch}.

\section{Background on Metric Learning} \label{metriclearning}

Metric learning is generally formulated as an optimization problem where one seeks to find the parameters of a distance function that minimize some objective function over the input data.
All algorithms currently implemented in \texttt{metric-learn} learn so-called Mahalanobis distances. Given a real-valued parameter matrix $L$ of shape \texttt{(n\_components, n\_features)} where \texttt{n\_features} is the
number of features describing the data, the associated Mahalanobis distance between two points $x$ and $x'$ is defined as $D_L(x, x') = \sqrt{(Lx-Lx')^\top(Lx-Lx')}$.
This is equivalent to Euclidean distance after linear transformation of the feature space defined by $L$.
Thus, if $L$ is the identity matrix, standard Euclidean distance is recovered.
Mahalanobis distance metric learning can thus be seen as learning a new
embedding space, with potentially reduced dimension \texttt{n\_components}.
Note that $D_L$ can also be written as $D_L(x, x') = \sqrt{(x - x')^\top M (x - x')}$, where we refer to $M = L^\top L$ as the Mahalanobis matrix.

\begin{figure}[t]
    \centering
    \begin{subfigure}[t]{0.12\textwidth} 
        \centering \includegraphics[scale=0.35]{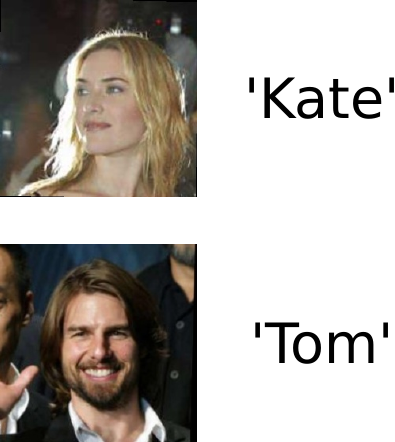}
        \caption{classes}\label{fig:full}
    \end{subfigure}
    \begin{subfigure}[t]{0.20\textwidth}
        \centering \includegraphics[scale=0.35]{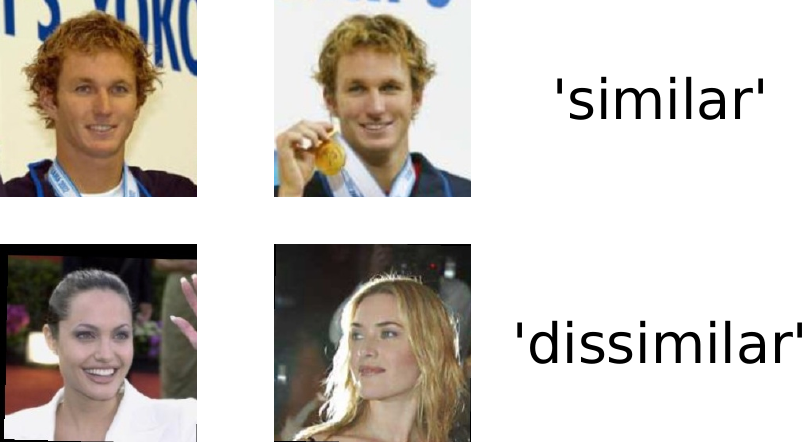}
        \caption{pairs}\label{fig:pairs}
    \end{subfigure}
    \begin{subfigure}[t]{0.31\textwidth}
        \centering \includegraphics[scale=0.35]{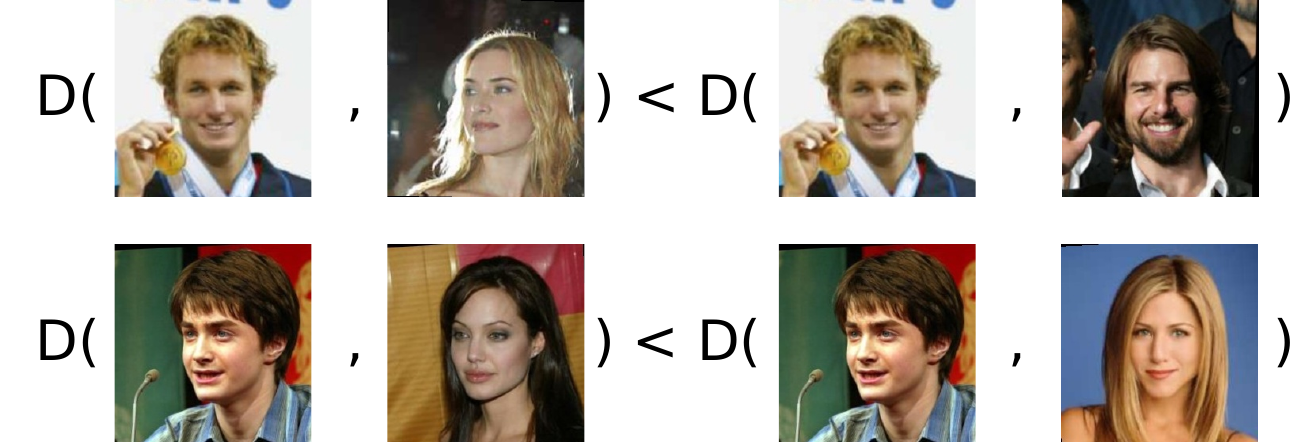}
	\caption{triplets}\label{fig:triplets}
    \end{subfigure}
    \begin{subfigure}[t]{0.31\textwidth}
        \centering \includegraphics[scale=0.35]{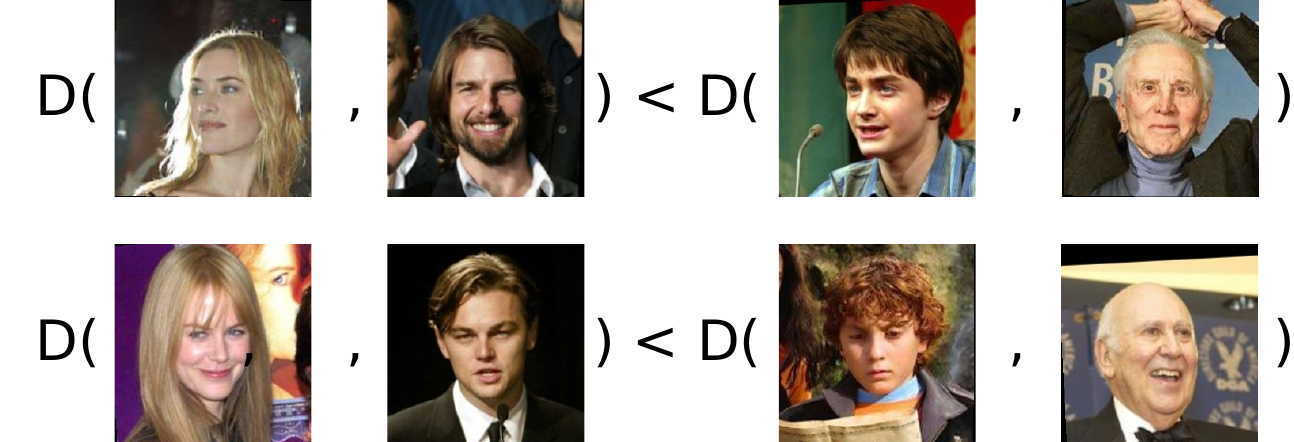}
        \caption{quadruplets}\label{fig:quadruplets}
    \end{subfigure}
    \caption{Different types of supervision for metric learning
    illustrated on face image data taken from the Labeled Faces in the Wild data set \citep{Huang12}.}\label{fig:flowers}
\end{figure}

Metric learning algorithms can be categorized according to the form of data supervision they require to learn a metric.
\texttt{metric-learn} currently implements algorithms that fall into the following categories.
\emph{Supervised learners} learn from a data set with one label per training example, aiming to bring together points from the same class while spreading points from different classes.
For instance, data points could be face images and the class could be the identity of the person (see Figure~\ref{fig:full}). 
\emph{Pair learners} require a set of pairs of points, with each pair labeled to indicate whether the two points are similar or not.
These methods aim to learn a metric that brings pairs of similar points closer together and pushes pairs of dissimilar points further away from each other.
Such supervision is often simpler to collect than class labels in applications when there are many labels.
For instance, a human annotator can often quickly decide whether two face images correspond to the same person (Figure~\ref{fig:pairs}) while matching a face to its identity among many possible people may be difficult. 
\emph{Triplet learners} consider 3-tuples of points and learn a metric that brings the first (\emph{anchor}) point of each triplet closer to the second point than to the third one.
Finally, \emph{quadruplet learners} consider 4-tuples of points and aim to learn a metric that brings the two first points of each quadruplet closer than the two last points.
Both triplet and quadruplets learners can be used to learn a metric space where closer points are more similar with respect to an attribute of interest, in particular when this attribute is continuous and/or difficult to annotate accurately (e.g., the hair color of a person on an image, see Figure \ref{fig:triplets}, or the age of a person, see Figure \ref{fig:quadruplets}).
Triplet and quadruplet supervision can also be used in problems with a class hierarchy.

\section{Overview of the Package}

The current release of \texttt{metric-learn} (v0.6.2) can be installed from the Python Package Index (PyPI) and conda-forge, for Python 3.6 or later.\footnote{Support for Python 2.7 and 3.5 was dropped in v0.6.0.}
The source code is available on GitHub at \url{http://github.com/scikit-learn-contrib/metric-learn} and is free to use, provided under the MIT license. 
\texttt{metric-learn} depends on core libraries from the SciPy ecosystem: \texttt{numpy}, \texttt{scipy}, and \texttt{scikit-learn}.
Detailed documentation (including installation guidelines, the description of the algorithms and the API, as well as examples) is available at \url{http://contrib.scikit-learn.org/metric-learn}.
The development is collaborative and open to all contributors through the usual GitHub workflow of issues and pull requests.
Community interest for the package has been demonstrated by its recent inclusion in the \texttt{scikit-learn-contrib} organization which hosts high-quality \texttt{scikit-learn}-compatible projects,\footnote{\url{https://github.com/scikit-learn-contrib/scikit-learn-contrib}} and by its more than 1000 stars and 200 forks on GitHub at the time of writing.
The quality of the code is ensured by a thorough test coverage (97\% as of June 2020).
Every new contribution is automatically checked by a continuous integration platform to enforce sufficient test coverage as well as syntax formatting with \texttt{flake8}.

Currently, \texttt{metric-learn} implements 10 popular metric learning algorithms.
Supervised learners include Neighborhood Components Analysis \citep[NCA,][]{Goldberger04}, Large Margin Nearest Neighbors \citep[LMNN,][]{Weinberger09}, Relative Components Analysis \citep[RCA,][]{Shental02},\footnote{RCA takes as input slightly weaker supervision in the form of \emph{chunklets} (groups of points of same class).} Local Fisher Discriminant Analysis \citep[LFDA,][]{Sugiyama07} and Metric Learning for Kernel Regression \citep[MLKR,][]{Weinberger07}.
The latter is designed for regression problems with continuous labels.
Pair learners include Mahalanobis Metric for Clustering \citep[MMC,][]{Xing2002a}, Information Theoretic Metric Learning \citep[ITML,][]{Davis07} and Sparse High-Dimensional Metric Learning \citep[SDML,][]{Qi09}.
Finally, the package implements one triplet learner and one quadruplet learner: Sparse Compositional Metric Learning \citep[SCML,][]{Shi15} and Metric Learning from Relative Comparisons by Minimizing Squared Residual \citep[LSML,][]{Liu12}.
Detailed descriptions of these algorithms can be found in the package documentation.

\section{Software Architecture and API}

\texttt{metric-learn} provides a unified interface to all metric learning algorithms.
It is designed to be fully compatible with the functionality of \texttt{scikit-learn}.
All metric learners inherit from an abstract \texttt{BaseMetricLearner} class, which itself inherits from \texttt{scikit-learn}'s \texttt{BaseEstimator}. All classes inheriting from \texttt{BaseMetricLearner} should implement two methods: \texttt{get\_metric} (returning a function that computes the distance, which can be plugged into \texttt{scikit-learn} estimators like \texttt{KMeansClustering}) and \texttt{score\_pairs} (returning the distances between a set of pairs of points passed as a 3D array).
Mahalanobis distance learning algorithms also inherit from a \texttt{MahalanobisMixin} interface, which has an attribute \texttt{components\_} corresponding to the transformation matrix $L$ of the Mahalanobis distance. \texttt{MahalanobisMixin} implements \texttt{get\_metric} and \texttt{score\_pairs} accordingly as well as a few additional methods. In particular, \texttt{transform} allows to transform data using \texttt{components\_}, and \texttt{get\_mahalanobis\_matrix} returns the Mahalanobis matrix $M=L^TL$.

Supervised metric learners inherit from \texttt{scikit-learn}'s base class \texttt{TransformerMixin}, the same base class used by \texttt{sklearn.LinearDiscriminantAnalysis} and others.
As such, they are compatible for pipelining with other estimators via \texttt{sklearn.pipeline.Pipeline}.
To illustrate, the following code snippet trains a \texttt{Pipeline} composed of LMNN followed by a k-nearest neighbors classifier on the UCI Wine data set, with the hyperparameters selected with a grid-search. Any other supervised metric learner can be used in place of LMNN.
\begin{minted}[%linenos=TRUE,
fontsize=\footnotesize]{python}
from sklearn.datasets import load_wine
from sklearn.neighbors import KNeighborsClassifier
from sklearn.model_selection import train_test_split, GridSearchCV
from sklearn.pipeline import Pipeline
from metric_learn import LMNN
X_train, X_test, y_train, y_test = train_test_split(*load_wine(return_X_y=True))
lmnn_knn = Pipeline(steps=[('lmnn', LMNN()), ('knn', KNeighborsClassifier())])
parameters = {'lmnn__k':[1, 2], 'knn__n_neighbors':[1, 2]}
grid_lmnn_knn = GridSearchCV(lmnn_knn, parameters, cv=3, n_jobs=-1, verbose=True)
grid_lmnn_knn.fit(X_train, y_train)
grid_lmnn_knn.score(X_test, y_test)
\end{minted}

Weakly supervised algorithms (pair, triplet and quadruplet learners) \texttt{fit} and \texttt{predict} on a set of tuples passed as a 3-dimensional array. Tuples can be pairs, triplets, or quadruplets depending on the algorithm.
Pair learners take as input an array-like \texttt{pairs} of shape \texttt{(n\_pairs, 2, n\_features)}, as well as an array-like \texttt{y\_pairs} of shape \texttt{(n\_pairs,)} giving labels (similar or dissimilar) for each pair.
In order to \texttt{predict} the labels of new pairs, one needs to set a threshold on the distance value.
This threshold can be set manually or automatically calibrated (at fit time or afterwards on a validation set) to optimize a given score such as accuracy or F1-score using the method \texttt{calibrate\_threshold}.
Triplet learners work on array-like of shape \texttt{(n\_triplets, 3, n\_features)}, where for each triplet we want the first element to be closer to the second than to the third one.
Quadruplet learners work on array-like of shape \texttt{(n\_quadruplets, 4, n\_features)}, where for each quadruplet we want the two first elements to be closer together than the two last ones. Both triplet and quadruplet learners can naturally \texttt{predict} whether a new triplet/quadruplet is in the right order by comparing the two pairwise distances.
To illustrate the weakly-supervised learning API, the following code snippet computes cross validation scores for MMC on pairs from Labeled Faces in the Wild \citep{Huang12}. Thanks to our unified interface, MMC can be switched for another pair learner without changing the rest of the code below.
\begin{minted}[%linenos=TRUE,
fontsize=\footnotesize]{python}
from sklearn.datasets import fetch_lfw_pairs 
from sklearn.model_selection import cross_validate, train_test_split 
from metric_learn import MMC
ds = fetch_lfw_pairs()
pairs = ds.pairs.reshape(*ds.pairs.shape[:2], -1)  # we transform 2D images into 1D vectors
y_pairs = 2 * ds.target - 1  # we need the labels to be in {+1, -1}
pairs, _, y_pairs, _ = train_test_split(pairs, y_pairs)
cross_validate(MMC(diagonal=True), pairs, y_pairs, scoring='roc_auc', 
               return_train_score=True, cv=3, n_jobs=-1, verbose=True) 
\end{minted}

\section{Future Work}

\texttt{metric-learn} is under active development. We list here some promising directions to further improve the package. To scale to large data sets, we would like to implement stochastic solvers (SGD and its variants), forming batches of tuples on the fly to avoid loading all data in memory at once.
We also plan to incorporate recent algorithms that provide added value to the package, such as those that can deal with multi-label \citep{liu15} and high-dimensional problems \citep{Liu19}, or learn other forms of metrics like bilinear similarities, nonlinear and local metrics \citep[see][for a survey]{Bellet15}.

\acks
We are thankful to Inria for funding 2 years of development. We also thank \texttt{scikit-learn} developers from the Inria Parietal team (in particular Gaël Varoquaux, Alexandre Gramfort and Olivier Grisel) for fruitful discussions on the design of the API and funding to attend SciPy 2019, as well as \texttt{scikit-learn-contrib} reviewers for their valuable feedback.

\bibliography{metric-learn-jmlr}

\end{document}